%% file: main.tex
\documentclass{article}
\usepackage{iclr2025_conference,times}
\usepackage{tabularx}
\usepackage{hyperref}
\usepackage{url}
\usepackage{graphicx} 
\usepackage{multirow}
\usepackage{booktabs}
\usepackage{rotating}
\usepackage{colortbl}
\usepackage{xcolor}
\usepackage{amsmath}
\usepackage{wrapfig}
\usepackage{multicol}
\usepackage{amssymb}
\usepackage{tcolorbox}
\usepackage{float}
\usepackage{tablefootnote}

\usepackage{longtable}     
\usepackage{caption}   
\usepackage{url}         
\usepackage{geometry}
\title{Granite Vision: a lightweight, open-source \\ multimodal model for enterprise Intelligence }
\author{Granite Vision Team, IBM Research \\
See Contributions and Acknowledgments section for full authors list \\
\texttt{granite-inquiries@ibm.com} 
}

\date{September 2024}

\newcommand{\squarebullet}[1]{\textcolor{#1}{\textbf{$\blacksquare$}}} 
\definecolor{SkyBlue}{RGB}{173, 216, 230}
\definecolor{LightGreen}{RGB}{144, 238, 144}
\definecolor{Peach}{RGB}{255, 178, 127}
\definecolor{Cyan40}{RGB}{15,98,254}
\definecolor{Cyan40}{RGB}{51,177,255}
\definecolor{Purple50}{RGB}{165,110,255}
\definecolor{Teal20}{RGB}{8,189,186}
\definecolor{Green40}{RGB}{56,183,98}

\definecolor{green}{RGB}{0. 205, 0}

\iclrfinalcopy 

\begin{document}

\maketitle

\begin{abstract}
We introduce Granite Vision, a lightweight large language model with vision capabilities, specifically designed to excel in enterprise use cases, particularly in visual document understanding. Our model is trained on a comprehensive instruction-following dataset, including document-related tasks, such as content extraction from tables, charts, diagrams, sketches, and infographics, as well as general image tasks. The architecture of Granite Vision is centered around visual modality alignment with a decoder-only, 2 billion-parameter Granite large language model. Additionally, we introduce a dedicated safety classification approach in test-time that leverages a sparse set of attention vectors to identify potential harmful inputs. Despite its lightweight architecture, Granite Vision achieves strong results in standard benchmarks related to visual document understanding, as well as on the LiveXiv benchmark, which is designed to avoid test set contamination by using a constantly updated corpus of recently published Arxiv papers. We are releasing the model under the Apache-2 license, allowing for both research and commercial use, while offering complete visibility into the training data and other relevant details. See \url{https://huggingface.co/ibm-granite/} for model weights.
\end{abstract}

\section{Introduction}

The confluence of computer vision and natural language processing has led to significant advances in multimodal learning, enabling large language models to effectively integrate and reason about visual content and linguistic data. Both proprietary and open-source multimodal models \citep{grattafiori2024llama3,geminiteam2024gemini15,OpenAI2023GPT4TR,abdin2024phi3technicalreporthighly,dai2024nvlm,zhang2024mm15,wang2024qwen2,agrawal2024pixtral,Li2024LLaVAOneVisionEV,chen2024internvl,molmo} have achieved remarkable success on a range of standard benchmarks, including visual question-answering, as well as more complex tasks such as multi-image reasoning and video understanding.

While this progress is exciting, a notable disparity remains: smaller models, typically with 1-4 billion parameters, have consistently fallen short of achieving  performance comparable to their larger counterparts containing tens of billions of parameters. In addition, existing models are predominantly trained on natural images, which can limit their performance in other domains, such as visual document understanding, where the unique visual characteristics, such as layouts, fonts, and graphics, significantly differ from natural images and require a more fine-grained comprehension of the visual content.

\begin{figure}[htpb]
 \centering
  \includegraphics[width=0.92\textwidth]{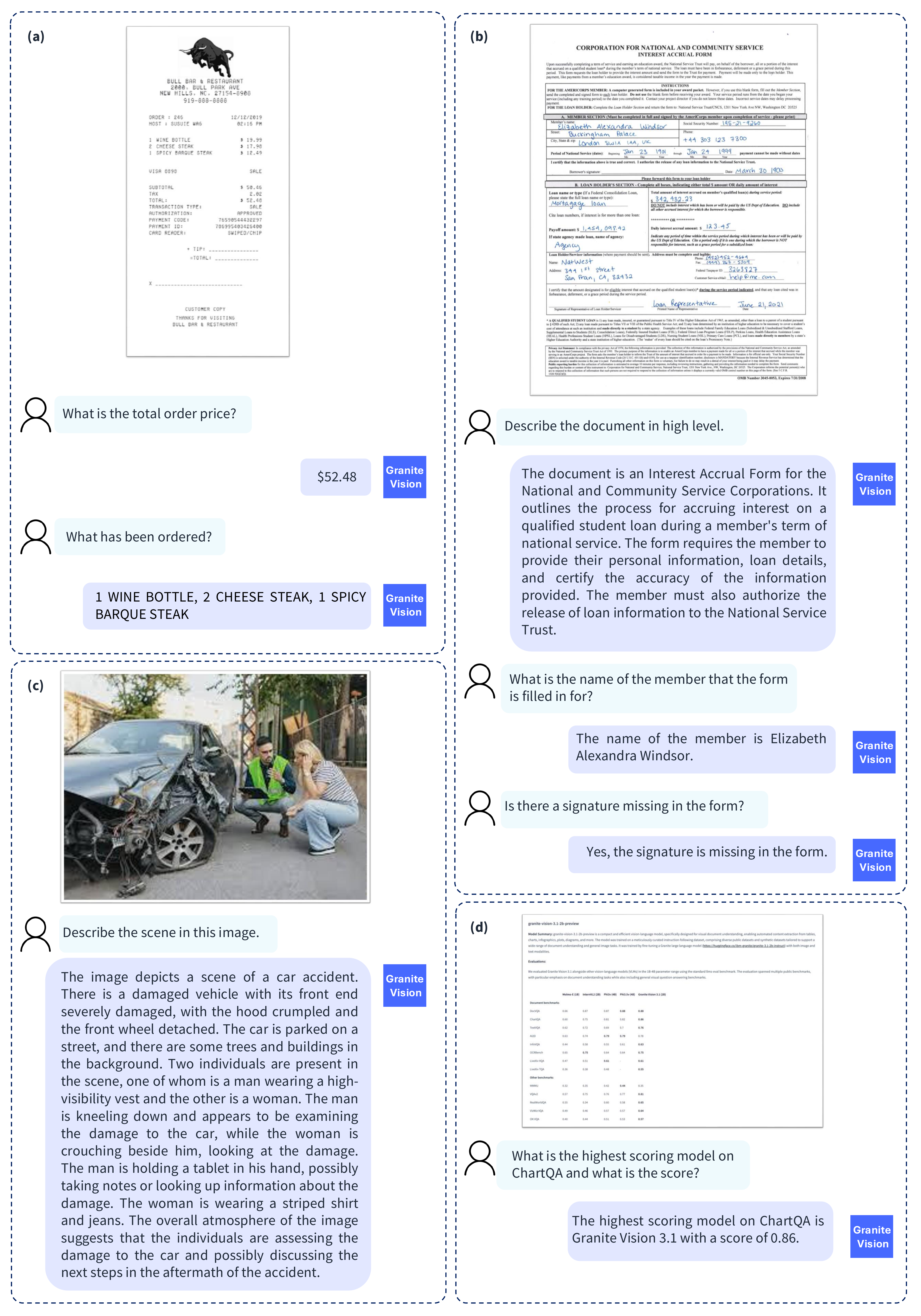}
 \caption{Qualitative examples generated by Granite Vision, showcasing its diverse capabilities including (a) document understanding such as receipt calculation, (b) form understanding with human handwritten text, (c) knowledge-grounded image description, (d) table understanding, etc. }
 \label{fig:qualitative}
 \end{figure}

In this work, we introduce Granite Vision, a compact vision-language model with approximately 3 billion parameters\footnote{We denote our model as Granite-Vision-3.1-2B, where the version (3.1) and size (2B) of the base large language model are explicitly indicated. However, when considering the integrated vision encoder and projector, the total parameter count of our model increases to 3 billion parameters.}, tailored to excel in enterprise use cases. Although our model can process general images, the first version of Granite Vision is particularly focused on visual document understanding, enabling automated content extraction from tables, charts, infographics, plots, diagrams, sketches, and more. Figure \ref{fig:qualitative} shows qualitative examples of our model’s output for visual document understanding tasks and general image description. Granite Vision extends the Granite family of large language models (~\cite{granite2024granite}), which have been trained on more than 12 trillion tokens, achieving state-of-the-art performance for their size, while being designed for enterprise usage, with full visibility into the training data.

As a key contribution of our work, we meticulously curate a comprehensive instruction-following dataset for visual document understanding, comprising around 13 million images and 80 million instructions, which span a diverse set of of tasks, including document question-answering , scene text understanding, key-value extraction, text grounding, layout parsing, captioning, UI understanding, and code (see Figure \ref{fig:doc_data}). We constructed this dataset by (i) mining documents from the web and synthesizing instructions as described in Section \ref{sec:docfm}, and (ii) unifying multiple public datasets into a single format. In addition to documents, our training data is further enriched by the inclusion of instruction-following data for general images from public datasets \citep{tong2024cambrian1,Li2024LLaVAOneVisionEV,laurencon2024matters,liu2023improvedllava}.

Similar to LLaVA \citep{liu2023llava}, our approach uses a projector to establish a connection between the visual encoder and a Granite large language model. Our training protocol consists of multiple stages, progressing from training the projector in isolation to jointly fine-tuning both the projector and the large language model, using denser image resolution grids at the latest stages. We also extract multi-layer features from the encoder, allowing us to better capture fine-grained details that are important for visual document understanding. Finally, we propose a novel method for safety classification as a separate module to identify harmful inputs based on a sparse set of attention vectors, which are selectively chosen to maximize safety classification performance.

Compared to other models of similar parameter size, Granite Vision achieves state-of-the-art results on established  benchmarks for visual document understanding, conversion of tables and charts to structured formats, and the LiveXiv benchmark \citep{shabtay2024livexiv}, which evaluates the model on recently published Arxiv papers, helping to mitigate the risk of test set contamination that can arise when models are trained on web-scraped data. In addition to its strong performance in enterprise settings, our model demonstrates competitive results on standard vision-language benchmarks. To promote openness and collaboration, we make the model publicly available under the Apache-2 license, allowing visibility into the training data and procedures.

\section{Related Work}

\subsection{Multimodal Large Language Models}

Recent advances in multimodal large language models (MLLMs) have demonstrated significant progress in understanding and generating content across different data modalities. Comprehensive surveys by \cite{Yin_2024_survey} and \cite{wadekar2024evolutionmultimodalmodelarchitectures} closely outline the various approaches to architecture design, training strategies, and evaluation methodologies in the field. 
The introduction of Flamingo \citep{alayrac2022flamingo} highlighted a remarkable performance of the transformer cross-attention architecture in vision-language tasks, serving as a catalyst for further advancements in the multimodal domain. The launch of GPT-4V \citep{openai_gpt4v} has sparked a competitive race in the development of powerful commercial-use MLLMs, leading to multimodal models like GPT-4o \citep{openai_gpt4o}, Claude 3.5 Sonnet \citep{anthropic_claude3_5}, and Gemini Pro 1.5 \citep{geminiteam2024gemini15}. In parallel, the open-source community has made significant contributions, with models like BLIP2 \citep{li2023blip2}, Phi-3.5-vision \citep{abdin2024phi}, LLaVA-OneVision \citep{li2024llavaonevision}, and Llama 3.2-vision \citep{grattafiori2024llama3} showing competitive performance on benchmarks while maintaining transparency and accessibility.
To mitigate the high costs associated with training large models in an end-to-end manner, a common approach involves employing modular architectures. These architectures typically combine a pre-trained modality encoder and a pre-trained large language model (LLM) with a learnable modality connector, with the latter comprising only a small portion of the total parameters \citep{bai2023qwenvl, tong2024cambrian1, chen2024internvl}. 
More recently, \cite{zhang2024mm15} demonstrated that even relatively small models can achieve strong performance with careful data curation and optimized training strategies.
Building on these insights, our model uniquely achieves state-of-the-art results on standard document and other benchmarks, all while operating at a significantly reduced scale (around 3 billion parameters). 

\subsection{Visual Document Understanding}

Visual document understanding, particularly the ability to comprehend charts, diagrams, tables, and document images, represents a crucial application area for MLLMs. 
Two primary technical challenges emerge in enabling MLLMs to process documents and associated images effectively: adequately encoding high-resolution images, and accurately interpreting visually-situated text within the documents. Recent approaches to addressing these challenges are often broadly categorized into two groups based on their text recognition methodology.
The first category, including models like DocFormer \citep{appalaraju2021docformer}, LayoutLMv3 \citep{huang2022layoutlmv3}, and, more recently, DocLLM \citep{wang2023docllm}, relies on external optical character recognition (OCR) systems to process text within images. The second category consists of ``OCR-free" models, e.g. Donut \citep{kim2022ocrfreedocumentunderstandingtransformer}, UReader \citep{ye2023ureader}, mPLUG-DocOwl 1.5 \citep{hu2024mplugdocowl15}, and DocPedia \citep{feng2024docpedia}.
Besides the specialized multimodal document understanding models \citep{liao2024doclayllm, liu2024hrvda}, strong performance is also achieved by general MLLMs instruction-tuned on document datasets \citep{li2024llavaonevision, dai2024nvlm}. 
While recent efforts \citep{mathew2021docvqa, huggingface_docmatix, rodriguez2024bigdocs} have produced several open-source document understanding datasets to advance model performance in this domain, large comprehensive datasets without restrictive licensing remain relatively limited.
Our approach builds upon this foundation by leveraging a comprehensive instruction-following dataset for visual document understanding, incorporating both synthetic data and public datasets.

\section{Data}\label{sec:data}
\begin{figure}[htpb]
    \centering
    \hspace{-20pt}
    \begin{minipage}{0.65\textwidth} 
        \centering
        \includegraphics[width=\textwidth]{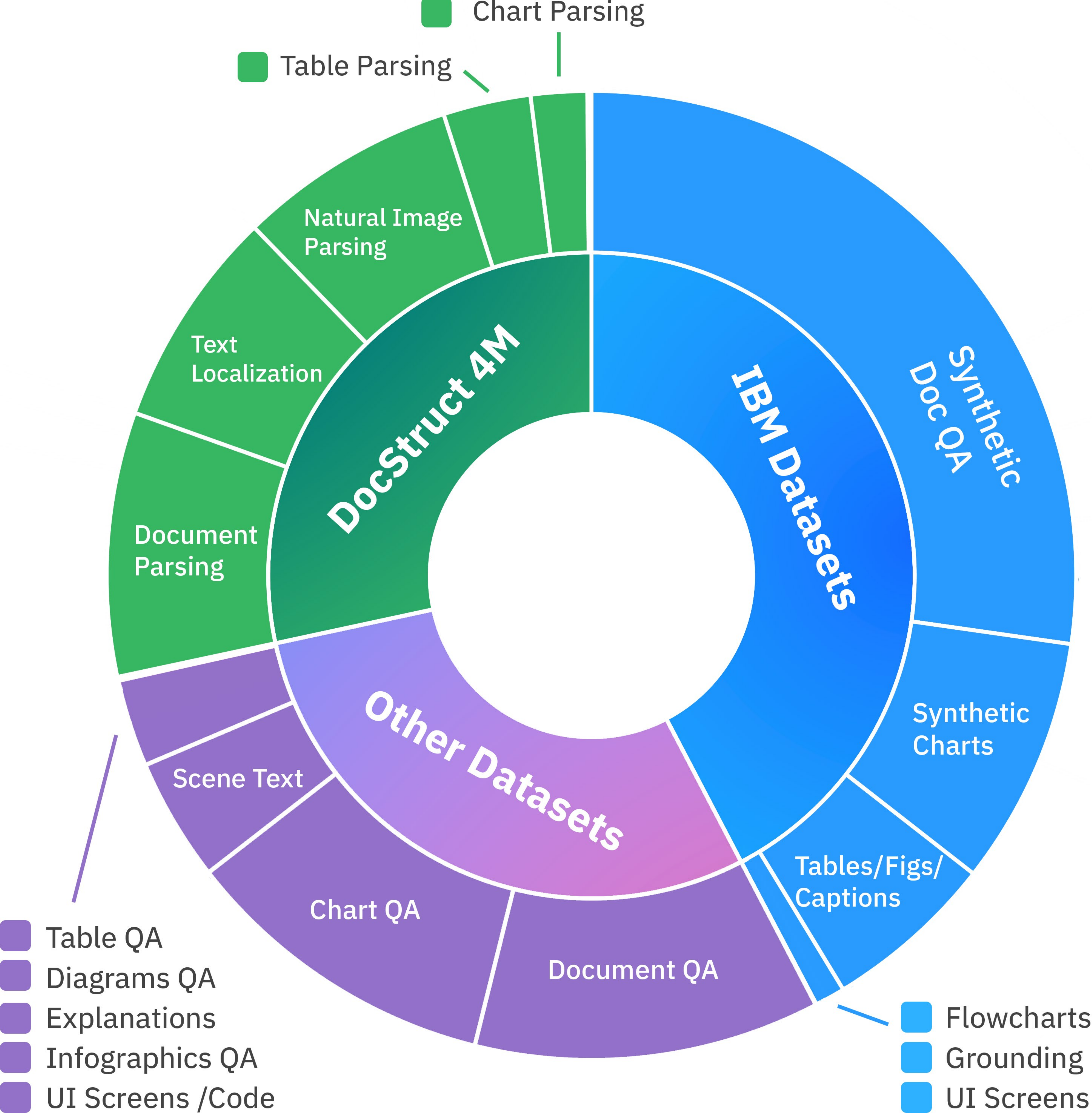} 
    \end{minipage}

    \vspace{6mm}
     
    \begin{minipage}{0.99\textwidth} 
        \centering
        \scriptsize
        \setlength{\columnsep}{0pt}
        \begin{multicols}{3} 
            \begin{itemize}
                \setlength{\itemsep}{0pt}
                
                \item[] \textbf{\fbox{Document QA}}
                \item[\squarebullet{Cyan40}] DocFM-VQA (2.4 M) 
                \item[\squarebullet{Purple50}] Docmatix (565.0 K) 
                \item[\squarebullet{Purple50}] synthdog-en (500.0 K) 
                \item[\squarebullet{Purple50}] pixmo-docs (255.4 K) 
                \item[\squarebullet{Purple50}] HME100K (74.5 K) 
                \item[\squarebullet{Purple50}] Ureader QA (72.7 K) 
                \item[\squarebullet{Purple50}] ArxivQA (54.4 K) 
                \item[\squarebullet{Purple50}] Ureader KG (37.6 K) 
                \item[\squarebullet{Cyan40}] Tech Book QA (10.4K)
                \item[\squarebullet{Purple50}] DocVQA (10.2 K) 
                \item[\squarebullet{Purple50}] IAM (5.7 K) 
                \item[\squarebullet{Cyan40}] Business Document QA (3.3K)
                \item[\squarebullet{Green40}] Visualmrc (3.0 K) 

                \item[] \textbf{\fbox{Chart QA}}
                \item[\squarebullet{Purple50}] Unichart (611.0 K) 
                \item[\squarebullet{Purple50}] Tinychart (606.0 K) 
                \item[\squarebullet{Cyan40}] DocFM-VQA-Charts (548.7 K) 
                \item[\squarebullet{Purple50}] Dvqa (200.0 K) 
                \item[\squarebullet{Green40}] PlotQA (157.0 K) 
                \item[\squarebullet{Green40}] FigureQA (100.0 K) 
                \item[\squarebullet{Cyan40}] DocFM-VQA-AugChartQA \hspace{-0.905mm}(35.1 K) 
                \item[\squarebullet{Purple50}] Chart2Text (27.0 K) 
                \item[\squarebullet{Green40}] ChartQA (18.3 K) 
                \item[\squarebullet{Purple50}] Vistext (9.9 K) 
                
                \vspace{1mm}
                
                \item[]\textbf{\fbox{Infographics QA}}
                \item[\squarebullet{Purple50}] InfographicVQA (8.5 K) 

                \item[] \textbf{\fbox{Table QA}}
                \item[\squarebullet{Purple50}] RoBUT-wikiSQL (75.0 K) 
                \item[\squarebullet{Purple50}] RoBUT-WTQ (38.0 K) 
                \item[\squarebullet{Purple50}] tabmwp (22.7 K) 
                \item[\squarebullet{Purple50}] RoBUT-SQA (8.5 K) 
                \item[\squarebullet{Purple50}] MultiHiertt (7.6 K)
                \item[\squarebullet{Purple50}] Finqa (5.3 K) 
                \item[\squarebullet{Purple50}] Hitab (2.5 K) 

                \item[] \textbf{\fbox{Diagram QA}}
                \item[\squarebullet{Purple50}] TQA (27.3 K) 
                \item[\squarebullet{Cyan40}] DocFM-VQA-Flowcharts (16.6 K) 
                \item[\squarebullet{Purple50}] AI2D GPT4V (4.9 K)
                \item[\squarebullet{Purple50}] AI2D InternVL (3.2 K)
                \item[\squarebullet{Purple50}] AI2D (2.4 K) 
                \item[\squarebullet{Purple50}] Diagram Image2Text (0.3 K)

                \item[] \textbf{\fbox{Reasoning / Captioning}}
                \item[\squarebullet{Cyan40}] DocFM Visual Cue \& Captioning \\(1.2M) 
                \item[\squarebullet{Purple50}] Textcaps (22.0 K) 
                \item[\squarebullet{Purple50}] DocReason (8.7 K) 

                \item[] \textbf{\fbox{Grounding / Text Localization}}
                \item[\squarebullet{Green40}] Multi-grained Text Localization \\(1.0 M) 
                \item[\squarebullet{Cyan40}] DocFM Rule Based Grounding \\(190 K) 
                \item[\squarebullet{Cyan40}] KVP10K (8.0 K) 
                
                \vspace{1mm}

                \item[] \textbf{\fbox{Scene Text}}
                \item[\squarebullet{Purple50}] K12 Printing (256.6 K) 
                \item[\squarebullet{Purple50}] OCR-VQA (165.7 K) 
                \item[\squarebullet{Purple50}] TextOCR (25.1 K) 
                \item[\squarebullet{Purple50}] Ureader CAP (22.0 K) 
                \item[\squarebullet{Purple50}] TextVQA (22.0 K) 
                \item[\squarebullet{Purple50}] Llavar (19.8 K) 
                \item[\squarebullet{Purple50}] ST-VQA (16.4 K) 
                \item[\squarebullet{Purple50}] RenderedText (10.0 K) 
                \item[\squarebullet{Purple50}] IIIT5K (2.0 K) 

                \item[] \textbf{\fbox{UI Screen / Code}}
                \item[\squarebullet{Purple50}] Datikz-v2 (94.4 K) 
                \item[\squarebullet{Purple50}] Datikz (47.9 K) 
                \item[\squarebullet{Purple50}] screenqa (33.2 K) 
                \item[\squarebullet{Purple50}] screen2words (15.7 K) 
                
                \item[\squarebullet{Purple50}] websight (10.0 K) 
                \item[\squarebullet{Purple50}] Omniact (4.9 K) 
                
                \item[] \textbf{\fbox{Structure Parsing}}
                \item[\squarebullet{Green40}] OCR-CC (1.0 M) 
                \item[\squarebullet{Green40}] CCpdf (937.8 K) 
                \item[\squarebullet{Purple50}] PubTables (569.3 K) 
                \item[\squarebullet{Cyan40}] DocFM-ChartExtraction (550.4 K)
                \item[\squarebullet{Green40}] TURL (200.0 K) 
                \item[\squarebullet{Green40}] PubTabNet (199.8 K) 
                \item[\squarebullet{Green40}] RVL-CDIP (159.4 K) 
                \item[\squarebullet{Green40}] DUE (104.5 K) 
                \item[\squarebullet{Cyan40}] FinTabNet (98.9 K) 

            \end{itemize}
        \end{multicols}
    \end{minipage}
    \vspace{0.2in}
    \caption{Overview of our comprehensive collection of document understanding datasets used for Granite Vision training.}
    \label{fig:doc_data}
\end{figure}

\begin{figure}[htbp]
    \centering
    \begin{minipage}{0.39\textwidth} 
        \centering
        \includegraphics[width=\textwidth]{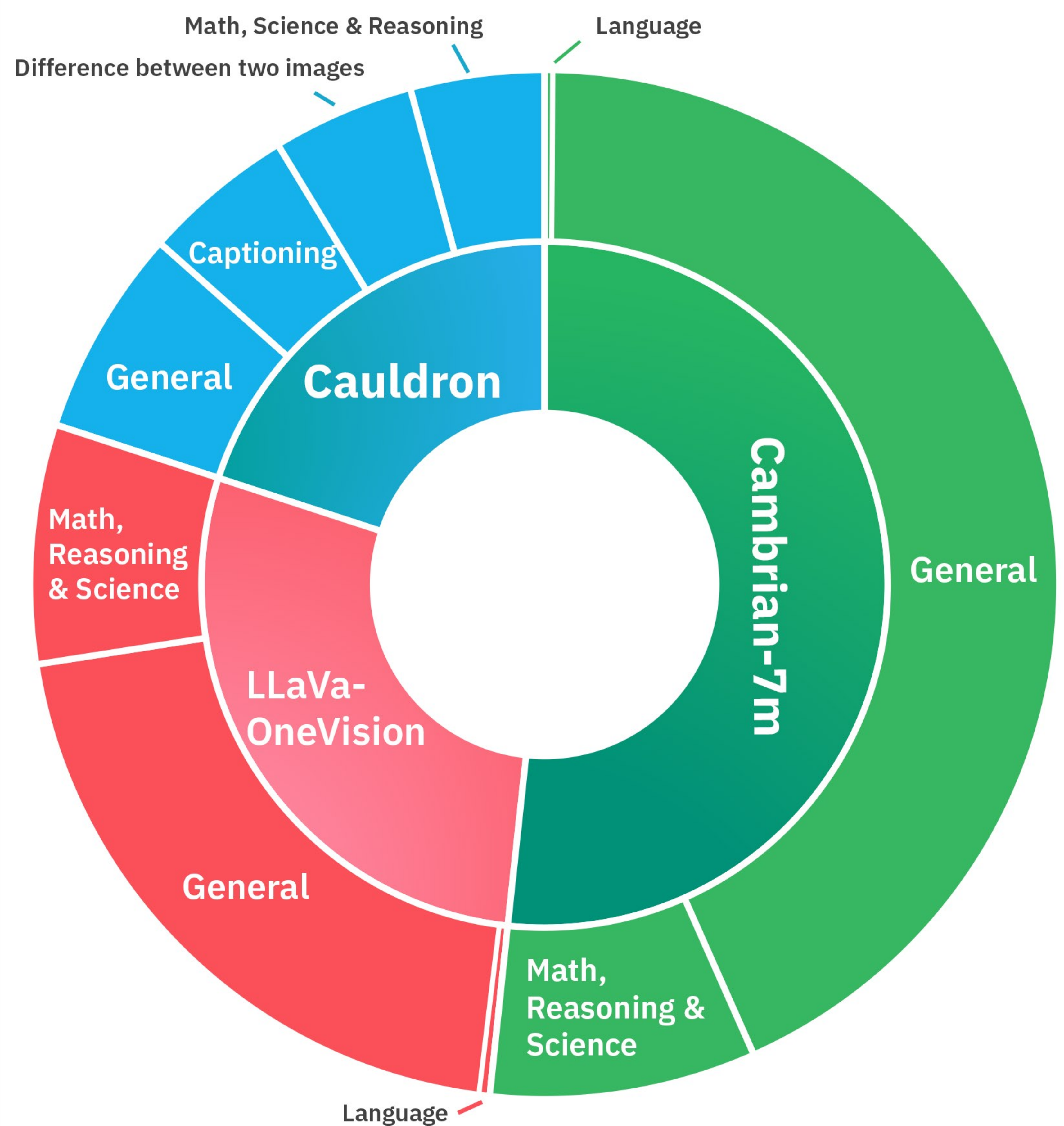} 
    \end{minipage}
    \hfill 
    \begin{minipage}{0.60\textwidth} 
        \centering
        \scriptsize
        \textbf{Image Count per Category Across Datasets} 
        \vspace{0.3cm}
        \begin{tabular}{l|ccc}
            \toprule
             & Cauldron & LLaVa-OneVision & Cambrian-7m \\
            \midrule
            General & 276.5 K & 881.3 K & 1.8 M \\
            Language/Captioning & 202.1 K & N/A & N/A\\
            Math/Science/Reasoning & 178.4 K & 318.0 K & 354.5 K\\
            Image Comparison & 188.9 K & N/A & N/A\\
            \bottomrule
        \end{tabular}
        
        \textbf{QA Count per Category Across Datasets} 
        \vspace{0.3cm}
        \begin{tabular}{l|ccc}
            \toprule
             & Cauldron & LLaVa-OneVision & Cambrian-7m \\
            \midrule
            General & 812.7 K & 2.0 M & 7.9 M \\
            Language/Captioning & 203.3 K & 1.2 M & 1.8 M\\
            Math/Science/Reasoning & 765.1 K & 464.8 K & 802.0 K\\
            Image Comparison & 237.9 K & N/A & N/A\\
            \bottomrule
        \end{tabular}
 
    \end{minipage}
    \caption{Overview of general image datasets used for Granite Vision training.}
    \label{fig:general_image_data}
\end{figure}

Granite Vision has been trained on a comprehensive instruction-following dataset, which covers a wide variety of visual categories ranging from document images to general images. The training data consists of a combination of pre-curated public vision datasets, common-crawl PDFs, and data that is synthesized in-house.
A detailed view of our data is presented in Figure \ref{fig:doc_data}, which depicts document understanding datasets, and Figure \ref{fig:general_image_data}, which represents general image datasets. 
Our document understanding data (Figure \ref{fig:doc_data}) covers a variety of document classes such as general document images, charts, flowcharts, diagrams and several more
encompassing a diverse set of visual Q\&A tasks. 
See Table \ref{tab:doc_datasets} (in Appendix) for a comprehensive overview of the datasets. Such diverse data is chosen to ensure the model’s ability to generalize across a wide array of document-centric applications. 
In addition to document datasets, we incorporated training data from multiple publicly available generic image datasets (Figure \ref{fig:general_image_data}). In the following sections, we dive deep into different data categories, data collection processes, synthetic data generation techniques, and data pre-processing pipelines. 

\subsection{IBM Curated Datasets} \label{sec:docfm}
In this section, we will describe several large synthetic document Q\&A datasets created at IBM and focused on a diverse set of document and Q\&A tasks. 
In order to seed this effort, we leveraged DocFM, a large-scale comprehensive dataset effort at IBM consisting of 85 million document pages extracted from unique PDF documents sourced from Common Crawl, Wikipedia, and ESG (Environmental, Social, and Governance) reports. DocFM served as a foundation for generating our synthetic datasets, which capitalized on its enriched nature for granular filtering and balancing.

DocFM was created using Docling \citep{DoclingTechReport}, an open-source document conversion toolkit developed by IBM. Docling facilitated the extraction and enrichment of each page with a rich set of attributes, resulting in a highly detailed representation of document pages.
Specifically, this enabled:
\begin{itemize}
    \item \textbf{Text and Positional Metadata Extraction:} Parsing each PDF page to retrieve textual content along with its positional metadata.
    \item \textbf{Layout Understanding:} Identification of structural components such as headers, footers, paragraphs, lists, captions, code, formulas, form elements, pictures, footnotes, and document indices.
    \item \textbf{Reading Order Detection:} Establishing the reading sequence of extracted content based on layout and parsed output.
    \item \textbf{Language Detection:} Identifying the primary language present on each page.
    \item \textbf{Figure Classification:} Labeling visual elements, such as charts, images, and diagrams.
\end{itemize}

To ensure high-quality textual representation, we focused exclusively on PDFs with extractable content, ensuring reliable and efficient text processing while minimizing noise. 

\subsubsection{DocFM - Data Collection \& Filtering}
A large portion of the DocFM dataset was obtained from the Common Crawl corpus which is an open repository of web data.
It is freely accessible and distributed as monthly archive files in various formats (``segments"). Among these segments, the URL index contains a list of crawled URLs accompanied by metadata, such as the media type specified by the HTTP header and the media type detected by Common Crawl from the actual URL content. By adhering to the Robots Exclusion Protocol~\citep{koster2022rfc} (commonly known as ``robots.txt"), Common Crawl ensures basic operational guarantees.
We obtained the Common Crawl dataset partition through the following process:
\begin{itemize}
\item A set of trusted web domains likely to contain relevant PDF files was manually curated. Additionally, websites of affiliations from public GitHub \footnote{https://github.com/} repositories, primarily universities and educational institutions, were identified. In total, we cataloged more than 45,000 domains to increase the relevance of the dataset and minimize potential issues related to web crawl data (e.g., toxic language, hate speech, abusive language, or profanity).
\item For each domain in the curated list, a search was conducted across the Common Crawl index segments of 13 archives (spanning May 2021 to January/February 2023) to identify URLs with the media type “application/pdf,” as either declared or detected.
\item For each discovered URL, we made an attempt to download the PDF file from the source. If successful, the file and its metadata were stored in a database.
\end{itemize}

Through this process, 4.96 million URLs were identified, resulting in 3.98 million PDF files after accounting for exact (binary) duplicates and invalid links. Additionally, we included 838,237 PDF documents retrieved from CCpdf~\citep{CCpdf}. 
After deduplicating, 4.6 million unique PDF files (75.5 million pages) remained in the Common Crawl partition. Following the conversion process, we successfully converted 65 million pages.

In addition, a total of 7,696 ESG reports were all sourced directly from \texttt{Responsibility Reports}\footnote{www.responsibilityreports.com}. To complement this, an automated pipeline using a headless Chrome browser component was employed to convert 20M pages of Wikipedia articles into PDF format. The list of articles was derived from the \texttt{wikipedia} dataset~\citep{wikidump}.


DocFM is a very large dataset which we plan to use for a variety of downstream visual learning applications in the future. For the current release of Granite Vision, we focused on a subset of DocFM for creation of a family of synthetic visual QA datasets (LLM and rule-based document and chart VQA) which will be described next.

\subsubsection{IBM Synthetic Visual Q\&A Datasets} 
\paragraph{DocFM-VQA (Synthetic Document VQA):}
There’s a lack of large-scale instruction-following datasets focused on visually-rich document understanding. The most commonly used dataset, DocVQA \citep{mathew2021docvqa}, is limited in size as it contains only 10K images and 40K QA pairs. 
Docmatix \citep{huggingface_docmatix} has been recently introduced as a potential large-scale alternative to DocVQA, but it uses only the textual parts of documents for QA pairs generation. 

In our work, we leverage a large language model (Mixtral 8x22B), to synthetically generate a large-scale visual question answering (VQA) dataset using a subset of DocFM as seed. We call this dataset DocFM verbalization-based VQA or DocFM-VQA in short. 
We employed a ``verbalized" representation of the documents as context for the LLM's (Mixtral 8x22B) QA generation. This involved extracting text from programmatic PDFs and augmenting it with verbal descriptions of visual elements such as charts, images, and tables. This approach encouraged the generation of questions that specifically target these elements. Our verbalization process included: 1) Converting charts to tables using DePlot \citep{liu-2022-deplot}, 2) Replacing images with captions generated by BLIP2 \citep{li2023blip2}. 3) Converting tables to markdown format using TableFormer \citep{tableformer}.
The DocFM-VQA contains a total of 20 million QA pairs from 2.4 million randomly selected pages from the DocFM dataset. This dataset facilitates the training of models capable of robust performance on visually rich documents.
While we explored the use of leading Visual Language Models (VLMs) to filter out potential hallucinations in the generated QAs, the observed performance benefits did not outweigh the computational cost. Therefore, filtering was not incorporated into the current model version.


\paragraph{DocFM-VQA-charts (Synthetic Chart VQA):}
A general framework for synthetic QA generation on charts is to use tabular data in textual form, using an LLM to generate QAs on it, and using code to generate the appropriate visual chart.
However, we observed two main issues with this approach.
First, the generated QAs are unable to refer to visual properties like the used markers or plot colors.
Second, LLMs tend to generate simple questions, such as a single data point extraction.
As a result, large-scale synthetic chart QA datasets tend to dramatically improve the performance on synthetically generated benchmarks but have a smaller effect on human questions or human-annotated benchmarks (e.g. ChartQA's augmented vs. human splits).

In our work, we addressed the two issues mentioned above by augmenting the tabular form of the data, used as input to the LLM for generating QAs, with additional information.
To solve the first issue of the LLM lacking knowledge of the visual appearance of the chart, we first choose a random marker and color for each data sequence. This information is added to the table consumed by the LLM and provided as input to the code used for rendering the chart.
To solve the second issue (of very simplistic QAs), we added to each table additional rows and columns. These rows and columns contain additional information calculated based on the original table using code. 
It includes max, min, sum, average, and a random operation (addition or subtraction) between two randomly chosen columns/rows, e.g. ``Column 1 $+$ Column 3".
This pre-calculated information encourages the model to ask questions requiring higher-order reasoning.
It also allows the model to produce more accurate ground-truth answers for cases that require arithmetic capabilities that LLMs struggle with \citep{schwartz-etal-2024-numerologic}.

We used the augmented table approach as input to the DocFM-VQA pipeline to generate the DocFM-VQA-Charts dataset. It contains 5.5M QAs on about 500K synthetic charts rendered based on random numeric tables extracted from the DocFM dataset.
We additionally augmented the ChartQA training set with the pre-calculated columns and rows information (without markers and colors) and generated additional QAs for the dataset's existing images.

\paragraph{DocFM-VQA-flowcharts (Synthetic Flowchart VQA):}
In order for the model to better understand flowcharts we generated synthetic flowcharts with QA pairs. We used an LLM (Mixtral 8x7B) for generating the flowchart nodes and edges. We then use those nodes and edges to query an LLM for QA pairs and to render a visual representation of the flowchart. We used the pipeline described below to generate 17K JSON representation and rendered flowcharts. The JSONs are fed to the DocFM-VQA pipeline to generate 76K QA pairs.

The flowchart synthesis pipeline:
\begin{itemize}
\item {\it Identify Business Domains and Industries}: Utilize the LLM to extract various business domains and identify industries associated with typical business processes.
\item {\it List Business Process Types}: For each business domain and industry, use the LLM to generate a comprehensive list of business process types.
\item {\it Generate Business Processes}: For each business process type, we employ the LLM to create multiple business processes constructed in a JSON format. Each process should have nodes and edges, with the number of nodes ranging from 5 to 20.
\item {\it Create Flowchart Images}: For each generated business process, use the ‘graphviz’ package to create a flowchart image. Incorporate random shapes, arrows, and colors to enhance visual diversity.
\end{itemize}

\paragraph{DocFM-ChartExtraction:}
\label{sec:chart_extraction}
We created a training dataset of 550K chart images to ensure that the model can perform structured data extraction from chart visualizations. The dataset was generated by converting tables extracted from DocFM and public stock market data into diverse chart types using matplotlib, including line, scatter, bar, pie, candlestick, and OHLC charts.
For each visualization, training examples included extraction instruction with randomized output format requirements (Markdown, HTML, or JSON) and varied natural language phrasings. To enhance document-level understanding capabilities, 100K DocFM-derived charts were embedded into their source document contexts by covering the original table with the chart image. The financial subset comprises 40K candlestick and OHLC charts generated from NASDAQ, S\&P500, and NYSE stock data.
This synthetic dataset enables the training of vision-language models to extract structured information from diverse visualization types while handling multiple output formats and document contexts.

\paragraph{DocFM Visual Cue and Captioning:} 
In order to create this dataset, we masked captions on images and then used the captions to generate questions regarding images. We implemented a multi-step process as follows: 1) Using Docling \citep{DoclingTechReport} we filtered to get figures associated with their corresponding captions; 2) We utilized Granite 3.1 8B model to analyze the captions for the presence of visual cues such as ``black arrow," ``on the left," ``red circle," and similar indicators of figure elements or regions of interest; 3) If the caption contained an explicit identifier (e.g., ``Figure 5"), we extracted it using the parsed text from the PDF. The caption body text was then masked on the image, leaving only the identifier visible. If no explicit identifier was present, we used the figure's bounding box as a fallback identifier. This identifier can then be used for the question, \textit{``What is the black arrow pointing at in Figure 5.?"}. 4) Leveraging the LLM, we generated context-specific questions about the figure, using the caption to inform the question formulation and answer while hiding the textual context in the image. 

\paragraph{DocFM Rule Based Grounding:}
We further complemented our data by implementing rule-based approaches to generate synthetic tasks focused on specific document elements like text localization, figures, and tables. Leveraging Docling's layout detector \citep{doclaynet}, we generated questions that query the location of text passages and other page elements. To encourage a deeper understanding of visual components, we masked captions associated with figures and tables, keeping only their identifiers. Subsequently, we employed templates to generate questions prompting the model to provide captions for these figures and tables. Furthermore, using defined templates we generated more operations such as selecting row, columns by index, or by value or comparison using only images. For the latter, we leverage the table understanding methods from Docling \citep{tableformer, otsl}.

\subsection{Public Datasets}
Our Granite Vision model is primarily focused on visual document understanding tasks. Therefore, we further enriched our data using a number of pre-curated high-quality, publicly available document-related datasets, which encompass several diverse tasks such as Document QA, Table QA, Chart QA, Diagram QA, OCR and scene-text related tasks, reasoning and grounding, UI screen/code and structure understanding. A detailed view of public document centric datasets is shown in Figure \ref{fig:doc_data}. For details readers are referred to Table \ref{tab:doc_datasets} (in Appendix). 



Beyond document-centric content, we also included high quality general image Q\&As sourced from public datasets 
to ensure robust performance in general visual tasks as well. Specifically, we acquired data from three high-quality collections: 
Cauldron\footnote{\url{https://huggingface.co/datasets/HuggingFaceM4/the_cauldron}}, Cambrian-7M\footnote{\url{https://huggingface.co/datasets/nyu-visionx/Cambrian-10M}} and LLaVa-OneVision-Data\footnote{\url{https://huggingface.co/datasets/lmms-lab/LLaVA-OneVision-Data}}. 
These datasets cover a wide range of domains and tasks, including image captioning, visual question answering (VQA), and object recognition, providing the foundational knowledge for general-purpose vision-language models. More details and statistics about general images data are shown in Figure ~\ref{fig:general_image_data}.

\subsection{Data Preprocessing}
Data pre-processing consisted of multiple annotation and filtering steps to conform to regulatory and safety requirements and improve overall data quality as described below.
\begin{itemize}
\item \textit{Restricted Data removal:}
All public datasets underwent a rigorous legal vetting process to identify data that could violate any legal or PII related requirements. All such data was removed, including datasets originating outside of the US and those with unclear license agreements.


\item \textit{Sexual Abuse Material (CSAM) removal:} All CSAM was removed from the datasets using state-of-the-art NSFW (not safe for work) detection methods. 

\item \textit{Face blurring:} Face blurring was performed to obfuscate all visual PII from the data. 

\item \textit{Deduplication:}
Since we combined multiple dataset collections, there could be duplication of text as well as visual data. 
Similar images were detected
through two methods: exact pixel-wise match, or perceptual hash matching \citep{zauner2010implementation}, for robustness against minor transformations.
Records with identical texts \textit{and} similar images were removed.
\end{itemize}

 
\section{Granite Vision Model}
\label{sec:model}

\begin{wrapfigure}{r}{0.5\textwidth} 
    \vspace{-0.5cm}
    \centering
    \includegraphics[width=0.5\textwidth]{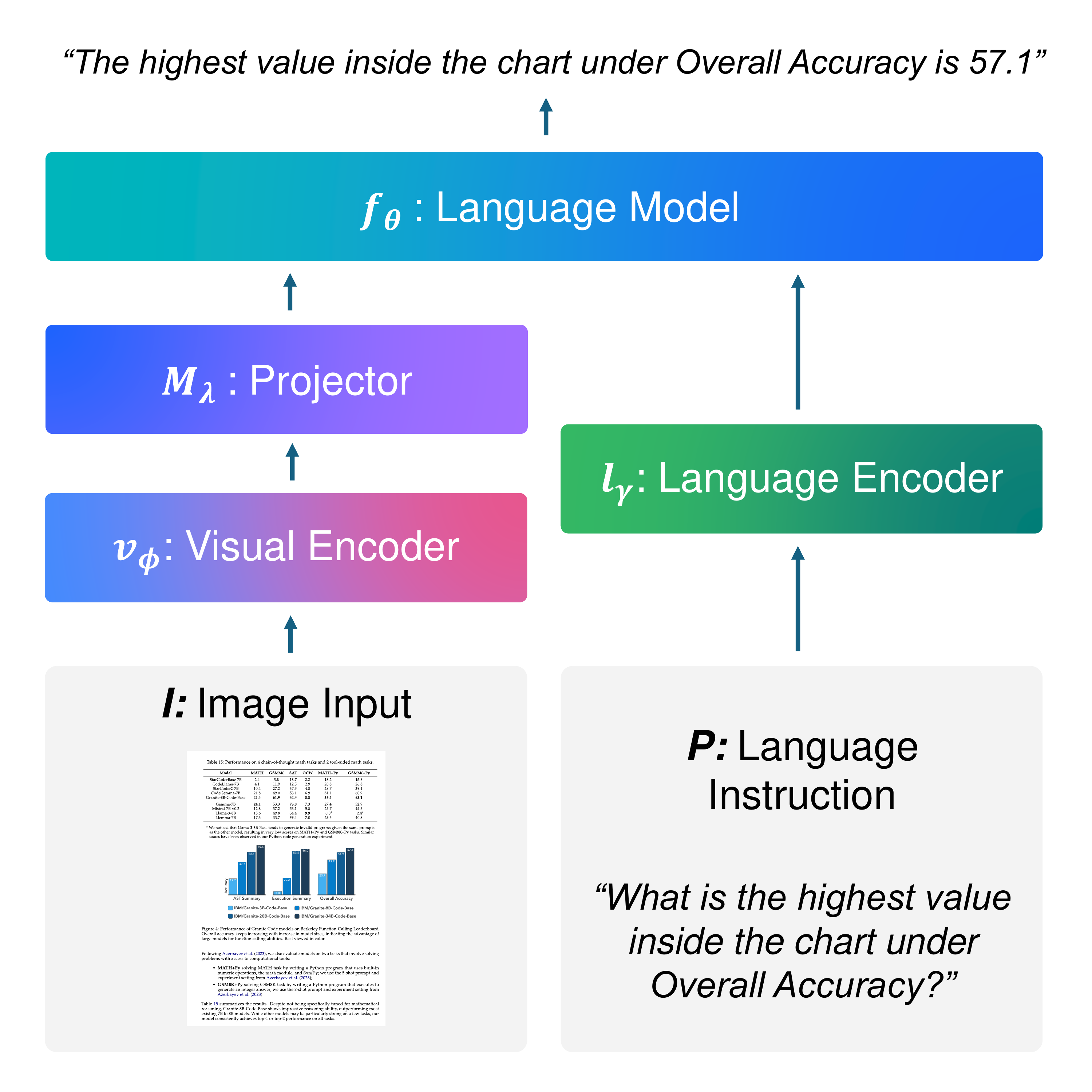}
    \vspace{-0.2cm}
    \caption{\textbf{Architecture of Granite Vision.}}
    \label{fig:architecture}
     \vspace{-0.2cm}
\end{wrapfigure}

In this section, we describe our model, including the architecture (Section~\ref{sec:model:arch}), the training procedure (Section~\ref{sec:model:training}), and the use of Sparse Attention Vectors (SAVs) for safety classification (Section~\ref{sec:model:sav}). An overview of our architecture is shown in Figure~\ref{fig:architecture}.

\subsection{Architecture}
\label{sec:model:arch}




Vision-and-language models are designed to process two data modalities simultaneously, such as images and corresponding text instructions. Each modality is encoded into a shared embedding space, which is then utilized for reasoning by a language model $f$ parameterized by $\theta$. Specifically, an image is encoded using a pre-trained visual encoder, denoted as $v$ parameterized by $\phi$, and a projector $M$ parameterized by $\lambda$. A corresponding textual prompt is tokenized and encoded using a fixed language encoder $l$ parameterized by $\gamma$. Given an input image $I$ and a text instruction $P$, the language model generates a text response $R$ as follows: $R = f_\theta(M_\lambda(v_\phi(I)), l_\gamma(P))$. This section provides a detailed description of each of these components.

\paragraph{Vision Encoder $v_\phi$.} For an input image $I$, we use a vision encoder $v_\phi$ to provide visual features $X=v_\phi(I)$. In our implementation, $X$ is a concatenation of outputs from multiple layers, allowing us to combine these different levels of representation, providing rich representations that are beneficial for visual document understanding. We use SigLIP~\citep{zhai2023siglip} as our vision encoder with $384 \times 384$ resolution. Additionally, we use the ``AnyRes'' technique in our training as in~\cite{li2024llavaonevision}, which is designed to accommodate images of various high resolutions. The ability to process high-resolution images is of particular importance for document understanding, especially when dealing with documents containing small text fonts, as it enables accurate text recognition.


Specifically, we use a global image, and a grid configuration that divides the image into multiple patches (or views), resulting in an additional multi-patch setting. We select the closest resolution from a predefined set that allows the original image size to be tiled into patches of size $384 \times 384$.
We tile the image using up to ten patches, leading to a wide variety of aspect ratios ranging from $1:10$ through $1:1$ to $10:1$ at different scales resulting in a total of 27 different tiling options. Next, the visual features $X$ are forwarded to the projector $M$.



\paragraph{Projector $M_\lambda$.} Here we apply $M_\lambda$, a two-layer MLP with GELU activation function as the projection layer to convert the visual features into a sequence of visual tokens: $T_{vis} = M_\lambda(X)$. This operation connects the image features into the word embedding space. We ensure the $T_{vis}$ tokens have the same dimensionality as the word embedding space
in the language model.


\paragraph{Language Encoder $l_\gamma$.} For a text instruction $P$, we extract the text tokens $T_{lang}$ as the instruction undergoes tokenization using a language encoder $T_{lang} = l_\gamma(P)$. 

\paragraph{Language Model $f_\theta$.} The visual tokens $T_{vis}$ and language tokens $T_{lang}$ are then concatenated and fed into a language model $f_\theta$. We use Granite 3.1-2B-Instruct~\citep{granite2024granite} as our language model $f_\theta$, which is trained for next token prediction. Granite 3.1 represents IBM’s third generation of large language models, released to the open-source community under an Apache 2.0 license, as well as matching top performance on general, enterprise, and safety language benchmarks. It supports up to 128K context length and shares a similar architecture as popular LLMs like Llama \citep{grattafiori2024llama3} to ensure compatibility with open-source inference and fine-tuning pipelines.


\subsection{Training Procedure}
\label{sec:model:training}


The training procedure of our Granite Vision model consists of three stages: (i) Pre-training for projector $M$, (ii)  Pre-training with language model $f_\theta$ and the Projector $M$, and (iii) Instruction tuning. Each image $I$ is accompanied by a text instruction $P$, and the predictions consist of the answer $A$ produced by the model. Specifically, for a response $R$, we compute the probability of the target words by the following equation:
\nolinebreak
\begin{equation}
p(A \mid I, P) = \prod_{i=1}^{|R|} p_{\psi}(x_i \mid I, P)
\end{equation}
\nolinebreak
where $\psi$ represents the trainable parameters. We note that $\psi$ contains the following trainable parameters mentioned above ($\theta, \phi, \lambda$), with different subsets are trained at various training stages. In addition, $x_i$ is the current prediction token. To calculate loss, we use the standard cross-entropy function with these probabilities. Next, we describe our three-step training process.

\paragraph{Stage 1: Pre-training for Projector $M_\lambda$.} 
In this step, the entire network (other than the projector $M_\lambda$) is frozen to correctly align visual and language tokens. For this stage, we use 558k image-text pairs with captions taken from LLaVA-Pretrain\footnote{\hyperlink{https://huggingface.co/datasets/liuhaotian/LLaVA-Pretrain}{https://huggingface.co/datasets/liuhaotian/LLaVA-Pretrain}.}. Additionally, we use only the following scales in this stage: (384, 768), (768, 384), (768, 768), (1152, 384), and (384, 1152).






Finally, we perform a hyperparameter search randomly on 5\% of the data to select a batch size of $512$, warm up ratio of $0.03$, and learning rate of $1e-4$ with a cosine scheduler. We also used the following multi-turn conversation template:

\begin{table*}[htp]\centering
\begin{minipage}{0.99\columnwidth}\vspace{0mm}    \centering
\label{multi-conversation}
\raggedright
\small
    ${\texttt{\color{green} <|system|>}}$ \\
    "A chat between a curious user and an artificial intelligence assistant. The assistant gives helpful, detailed, and polite answers to the user's questions." \\
     $\texttt{\color{green} <|user|>} P^{1} \quad \texttt{\color{green} <|assistant|>} A^{1} \texttt{\color{green}<|end\_of\_text|>}$\\
    $\texttt{\color{green} <|user|>} P^{2} \quad \texttt{\color{green} <|assistant|>} A^{2} \texttt{\color{green}<|end\_of\_text|>}$
\end{minipage}
\end{table*}

\noindent where ($P^i$, $A^i$) is the $i$-th conversation pair iteration.

\paragraph{Stage 2: Pre-training with Language Model $f_\theta$ for Projector $M$.} In order to align the LLM with the updated visual features that have been processed by the projector, we use a second pre-training stage, in which we keep the vision encoder frozen and fine-tune the LLM $f_\theta$ and the Projector $M_\lambda$. We use the same image-text pairs data and same multi-conversation template as in stage 1.



It is important to note that at the end of this stage, we retain only Projector $M_\lambda$ weights and discard the fine-tuned LLM, reverting the language model to its original weights for the next stage. We found that this procedure leads to a better starting point for the following instruction tuning stage.


In stage 2, for the LLM and Projector $M_\lambda$, we use different learning rates of $8e-05$ and $32e-05$, respectively, and kept the same other stage 1 hyperparameters.






\paragraph{Stage 3: Instruction Tuning.} In the last stage, we perform supervised fine-tuning with all the instruction-following data. Here, we update both the pre-trained weights of the projector $M$ and the language model $f_\theta$, while keeping the visual encoder weights frozen. We use approximately 20M image-text pairs from our collected datasets, as described in Section~\ref{sec:data}. We retained the same hyperparameters from stage 2, with the exception of the batch size, which was set to $1024$.



 
\subsection{Sparse Attention Vectors for Safety}
\label{sec:model:sav}


Here we outline how we use sparse attention vectors (as proposed in ~\cite{mitra2024sparse} and~\cite{Huang2024MultimodalTV}), derived from the activation space of a Granite Vision model, as features for a safety classification setup. Our key insight is that within Granite Vision's many attention heads and transformer layers, there is a sparse subset of features that are useful for identifying safety if we formalize a range of safety tasks as classification problems. We present the following three-step method to identify and utilize these safety features.


\paragraph{Step 1: Extracting General Attention Vectors.}
Given our model and few-shot samples of label pairs
$\{(x_1, y_1), (x_2, y_2), \dots, (x_N, y_N)\}$,
we first extract the attention vectors for each sequence $x_i$. Here, the labels in the few-shot demonstrations indicate specific safety characteristics of the query, like 'harmful' and 'non-harmful' classes. Next, for every $x_i$, we compute the attention vector $\mathbf{h}_l^m$ for head $m$ from layer $l$ for the final token $x_i^T$. This yields a set of attention vectors $\{\mathbf{h}_l^m(x_i^T) \mid i = 1, \dots, N\}$ for each head $m$ and layer $l$.

\paragraph{Step 2: Finding Safety Vectors.}
In order to identify which attention vectors are naturally suited to a safety task, we evaluate each vector's discriminative ability by computing its performance under a nearest class centroid classifier. 

Specifically, for each class $c \in \mathcal{C}$, we compute its mean attention vector across few shot examples:
$$\mu_c^{l,m} = \frac{1}{|N_c|}\sum_{c} \mathbf{h}_l^m(x_j^T)$$
where $N_c$ is the set of indices of examples with label $c$. For each input $x_i$, we compute its cosine similarity to each class centroid head:
$$s_{l,m}(x_i, c) = \frac{\mathbf{h}_l^m(x_i^T) \cdot \mu_c^{l,m}}{\|\mathbf{h}_l^m(x_i^T)\| \|\mu_c^{l,m}\|}, \quad \forall c \in \mathcal{C}$$

Next, we measure the discriminative ability of each head by its performance as follows:
$$
\text{score}(l, m) = \sum_{i=1}^{N} \mathbf{1}[\hat{y} = y_i] $$

\noindent where the nearest class centroid label is given as $\hat{y}$, and $\mathbf{1}[\cdot]$ is the indicator function that evaluates to 1 when the condition is true. We denote the set of $k$ top-scoring heads as $\mathcal{H}_\text{SAV}$:
$$
\mathcal{H}_\text{SAV} = \{(l,m) \mid \text{score} (l,m) \hspace{0.5em} \text{is} \hspace{0.5em}  \text{among} \hspace{0.5em} \text{top K} \}
$$

\paragraph{Step 3: Classification with Safety Vectors.}

Given a sample sequence input $Q$ to classify, we leverage our sparse set of heads $\mathcal{H}_\text{SAV}$ for safety prediction. For each head $(l,m) \in \mathcal{H}_\text{SAV}$, we compute the class centroid $\mu_c^{l,m}$ closest to the sequence input as follows:
$$
\hat{y}_{l,m} = \underset{c \in \mathcal{C}}{\arg\max}\ s_{l,m}(Q^T, c)
$$
where $s_{l,m}$ is defined as in Step 2.
The final prediction counts the majority across all heads in $\mathcal{H}_\text{SAV}$:
$$
\underset{y \in \mathcal{C}} {\arg\max} \sum_{(l,m) \in \mathcal{H}_\text{SAV}} \mathbf{1}[\hat{y}_{l,m} = y]
$$

Using this approach, we are able to determine whether a new input sequence is safe. We note that this approach can be used more generally, as it has been used in~\cite{mitra2024sparse} to extract multimodal features from large multimodal models for general discriminative vision-language tasks.


\section{Evaluation}

\subsection{Standard benchmarks}

We evaluated our models on a set of popular public benchmarks. Since the current Granite Vision release is mainly geared towards document understanding, our focus is on document-related benchmarks. However, we also report results for key natural-image benchmarks. 
We used the standardized \textit{lmms-eval} package \citep{zhang2024lmmsevalrealitycheckevaluation,lmms_eval2024} to run the evaluations. We do not use any test time optimizations such as prompt tuning or chain-of-thought.
The results for most other models were also produced by running the standard \textit{lmms-eval} benchmarks when possible.
For the remaining models, we used the results reported by the original authors or from the publicly reported benchmarks.

\paragraph{Document related benchmarks}
\begin{itemize} 
\item \textbf{DocVQA}: A benchmark designed to evaluate the models' ability to understand and extract textual information from documents \citep{mathew2021docvqa}.

\item \textbf{ChartQA}: A benchmark which focuses on question answering about charts, requiring both visual and logical reasoning to interpret data presented in various chart formats \citep{masry2022chartqa}.

\item \textbf{TextVQA}: Challenges models to read and reason about text present within images to answer questions accurately \citep{yang2021tap}.

\item \textbf{AI2D}: Contains grade school science diagrams, aimed at evaluating diagram understanding and question answering capabilities \citep{kembhavi2016diagram}.

\item \textbf{InfoVQA}: Contains infographics - documents that combine textual, graphical, and visual elements to communicate information effectively. The questions require models to perform joint reasoning across document layout, text, and data visualizations \citep{mathew2022infographicvqa}.

\item \textbf{OCRBench}: A benchmark designed to evaluate Optical Character Recognition (OCR) capabilities within various contexts, assessing the accuracy and robustness of OCR in documents and in-the-wild \citep{liu2024ocrbench}.

\item \textbf{WebSRC}: A Web-based Structural Reading Comprehension benchmark. It consists of screenshots and question-answer pairs created based on the corresponding HTML source code and metadata. Each question in WebSRC requires a certain structural understanding of a web page to answer, and the answer is either a text span on the web page or yes/no.

\item \textbf{LiveXiv}: A benchmark  which specifically focuses on testing multi-modal models' ability to understand domain-specific visual content like graphs, charts, and tables from scientific manuscripts. It is a live benchmark, containing recently published papers on Arxiv, helping prevent test set contamination that can occur with static benchmarks when models are trained on web-scraped data. It contains questions dealing with figures (VQA) and with tables (TQA) \citep{shabtay2024livexiv}.

 \end{itemize}

\paragraph{Natural image benchmarks}
\begin{itemize} 
\item \textbf{MMMU}: A comprehensive benchmark designed to evaluate Multimodal Large Language Models on both perception and cognition abilities across various subtasks, ensuring robust and diverse testing of these models \citep{yue2024mmmu}.

\item \textbf{VQAv2}: A benchmark for Visual Question Answering containing open-ended questions about images, designed to test a model's ability to understand and reason about visual content \citep{goyal2017making}.

\item \textbf{RealWorldQA}: Focuses on question answering in real-world scenarios, requiring models to interpret and reason over complex visual and textual information present in everyday images.

\item \textbf{VizWiz}: Contains over 31,000 visual questions originating from blind individuals, aiming to help answer visual questions posed by this community, with each question accompanied by crowdsourced answers \citep{gurari2018vizwiz}. 

\item \textbf{OK VQA}: A benchmark designed for visual question answering tasks that require external knowledge beyond what is visible in the image, featuring over 14,000 questions to evaluate the reasoning abilities of AI models \citep{marino2019ok}.
 \end{itemize}

The results are presented in Table \ref{tab:model_comparison}. Granite Vision demonstrates impressive performance across document-related and natural image benchmarks. In the small model category (1B-4B), it achieves leading scores on most key benchmarks, e.g. DocVQA (88\%) and ChartQA (86\%), outperforming other models. Particularly notable is that its performance on these tasks is competitive with much larger proprietary models, such as Gemini-1.5-Pro, GPT-4o, and GPT-4V, which have significantly more parameters. The results highlight that Granite-Vision's data curation and training approach are particularly effective for document understanding tasks, achieving strong results despite its compact size.

\begin{table}[t]
\centering
\setlength{\tabcolsep}{3pt}  
\begin{tabular}{lc|ccccccccc|cccccc}
\toprule
 & & \multicolumn{9}{c}{Document benchmarks} & \multicolumn{5}{c}{Other benchmarks} \\
Model & Size & \rotatebox{90}{DocVQA} & \rotatebox{90}{ChartQA} & \rotatebox{90}{TextVQA} & \rotatebox{90}{AI2D} &\rotatebox{90}{InfoVQA} & \rotatebox{90}{OCRBench} & \rotatebox{90}{WebSRC} &\rotatebox{90}{LiveXiv VQA}  &\rotatebox{90}{LiveXiv TQA}& \rotatebox{90}{MMMU}& \rotatebox{90}{VQAv2}& \rotatebox{90}{RealWorldQA}& 
\rotatebox{90}{VizWiz VQA}& \rotatebox{90}{OK VQA} \\
\midrule
\rowcolor{gray!20} \multicolumn{15}{l}{Small models 1B-4B} &\\
Molmo-E & 1B & 0.66 & 0.60 & 0.62 & 0.63 & 0.44 & \underline{0.65} & 0.68 & 0.47 & 0.36 & 0.32 & 0.57 & 0.55 & 0.49 & 0.40 \\
MM1.5* & 1B & 0.81 & 0.67 & \underline{0.72} & 0.59 & 0.50 & - & - & - & - & 0.36 & - & 0.53 & - & - \\
SmolVLM* & 2.2B & 0.80 & {0.72} & \underline{0.72} & \textbf{0.84} & - & \underline{0.65} & - & - & - & 0.38 & - & - & - & - \\
MM1.5* & 3B & \underline{0.87} & 0.74 & \textbf{0.76} & 0.66 & 0.58 & - & - & - & - & 0.37 & - & 0.57 & - & - \\
Phi3v & 4B & \underline{0.87} & 0.81 & 0.69 & \underline{0.79} & 0.55 & 0.64 & \textbf{0.91} & \underline{0.61} & 0.48 & \underline{0.42} & 0.76 & \underline{0.60} & \underline{0.57} & 0.51 \\
Phi3.5v & 4B & \textbf{0.88} & \underline{0.82} & 0.70 & \underline{0.79} &\underline{ 0.61} & 0.64 & \textbf{0.91} & \textbf{0.63} & \underline{0.51} & \textbf{0.44} & \underline{0.77} & 0.58 & \underline{0.57} & \underline{0.53} \\
\rowcolor{green!20} Granite Vision & 3B & \textbf{0.88} & \textbf{0.86} &\textbf{ 0.76} & 0.78 & \textbf{0.63} & \textbf{0.75} & \underline{0.90} & \underline{0.61} & \textbf{0.55} & 0.35 & \textbf{0.81} & \textbf{0.65} & \textbf{0.64} & \textbf{0.57} \\

\midrule
\rowcolor{gray!20} \multicolumn{15}{l}{Mid-size models 7B-13B} &\\
Molmo-D & 7B & 0.73 & 0.58 & 0.70 & 0.79 & 0.52 & 0.70 & 0.76 & 0.61 & 0.51 & - & 0.68 & 0.51 & 0.45 & 0.51 \\
Molmo-O & 7B & 0.70 & 0.59 & 0.59 & 0.77 & 0.50 & 0.66 & 0.77 & 0.60 & 0.46 & - & 0.38 & 0.67 & 0.56 & 0.39 \\
MM1.5* & 7B & 0.88 & 0.78 & 0.76 & 0.72 & 0.59 & - & - & - & - & 0.42 & - & 0.62 & - & - \\
Cambrian-1* & 8B & 0.78 & 0.73 & 0.72 & 0.73 & - & 0.62 & - & - & - & 0.43 & - & 0.64 & - & - \\
Llama3.2* & 11B & 0.88 & 0.83 & - & 0.91 & - & - & - & - & - & 0.51 & 0.75 & - & - & - \\
Pixtral & 12B & 0.91 & 0.83 & 0.76 & 0.80 & 0.56 & 0.66 & 0.90 & 0.75 & 0.61 & 0.50 & 0.79 & 0.63 & 0.58 & 0.59 \\
Cambrian-1* & 13B & 0.77 & 0.74 & 0.73 & 0.74 & - & 0.62 & - & - & - & 0.40 & - & 0.63 & - & - \\

\midrule
\rowcolor{gray!20} \multicolumn{15}{l}{Larger models} &\\
MM1.5* & 30B & 0.91 & 0.83 & 0.79 & 0.77 & 0.67 & - & - & - & - & 0.47 & - & 0.69 & - & - \\
Cambrian-1* & 34B & 0.76 & 0.76 & 0.77 & 0.80 & - & 0.60 & - & - & - & 0.50 & - & 0.68 & - & - \\
NNLM-D* & 72B & 0.92 & 0.86 & 0.82 & 0.94 & - & - & - & - & - & 0.60 & 0.85 & 0.70 & - & - \\
Llava-OV* & 72B & 0.93 & 0.84 & - & 0.86 & 0.79 & - & - & - & - & 0.57 & - & 0.72 & - & - \\
Llama3.2* & 90B & 0.90 & 0.85 & - & 0.92 & - & - & - & - & - & 0.60 & 0.78 & - & - & - \\
Claude-Sonnet* & - & 0.95 & 0.91 & - & 0.95 & - & - & - & 0.75 & 0.83 & 0.68 & - & - & - & - \\
Gemini-1.5-Pro* & - & 0.93 & 0.87 & 0.79 & 0.94 & 0.81 & - & - & - & - & 0.62 & 0.80 & 0.70 & - & - \\
GPT-4V* & - & 0.88 & 0.78 & - & 0.76 & - & - & - & - & - & 0.53 & - & 0.56 & - & - \\
GPT-4o* & - & 0.93 & 0.85 & - & 0.84 & - & - & - & 0.60 & 0.54 & 0.69 & - & 0.75 & - & - \\

\bottomrule
\end{tabular}
\caption{Performance comparison across different models and benchmarks. For models marked with * we report only the available numbers from the original publication or the published benchmark and did not run the evaluations in the controlled \textit{lmms-eval} setup.}
\label{tab:model_comparison}
\end{table}

\subsection{Additional evaluations}
Converting tables and charts to a structured format are important tasks for enterprise use cases. We added to lmms-eval the benchmarks described below and compared our model with other open models.

\paragraph{Table extraction}
To evaluate the ability of models to turn a table image into a structured format, we used the test sets of 
PubTables \citep{smock2022pubtables} and 
FinTabNet \citep{chen2021finqa}.
We instruct the model to extract the data in HTML table format
and evaluate performance according to Tree-Edit-Distance-based Similarity (TEDS) from \cite{zhong2020image}. It measures both the table structure similarity and each cell string edit distance.

\paragraph{Chart extraction}
Similar to table extraction evaluation, we also evaluate the ability of models to turn a chart image into a structured format. We used the test set of ChartQA \citep{masry2022chartqa} for this evaluation.
We instruct the model to extract the data in Markdown or HTML format, the prompt includes an example of the expected format.
Model performance is evaluated according to a Modified TEDS (mTEDS) metric. The original TEDS measures string edit distance for each cell, which is not optimal for numerical values where the distance should take into account the scale of chart values. We define mTEDS as follows, let $T$ be a table with numeric values $V=\{v_{ij}\}$ excluding headers. Define the normalized function:
$$
N(v_{ij}) = round\left(20 \cdot \frac{v_{ij}}{\max_{i,j}(abs(V_{gt}))}\right),$$
where $V_{gt}$ represents the ground truth table values. The metric applies standard TEDS computation on the normalized values, comparing structural and content similarity between predicted and ground truth tables. This normalization accounts for scale-dependent accuracy in chart data extraction while maintaining TEDS's ability to evaluate structural correctness.

The results are reported in Table \ref{tab:model_comparison_chart_extraction}.
Granite Vision significantly outperforms other models in its size category.
It shows an advantage of $+12\%$-$+39\%$ over the second-best models in tables and charts extraction to markdown and HTML.
Notably, Granite Vision's performance is comparable to the 4 times larger Pixtral model.

\begin{table}[t]
\centering
\begin{tabular}{lccccc}
\toprule
 &  & \multicolumn{2}{c}{Table Extraction} & \multicolumn{2}{c}{Chart Extraction} \\
Model & Size &  {PubTables} & FinTabNet & {MD} & HTML \\

\midrule
\rowcolor{gray!20} \multicolumn{6}{l}{Small models 1B-4B} \\
Molmo-E & 1B & 0.28 & 0.28 & 0.57 & 0.54 \\
SmolVLM & 2.2B & 0.32 & 0.18 & 0.12 & 0.02 \\
Phi3.5v & 4B & 0.58 & 0.28 & 0.77 & 0.40 \\
\rowcolor{green!20}
Granite Vision & 3B & \textbf{0.70} & \textbf{0.54} & \textbf{0.93} & \textbf{0.95} \\
\midrule
\rowcolor{gray!20} \multicolumn{6}{l}{Larger models 7B-12B} \\
Molmo-D & 7B & 0.34 & 0.25 & 0.62 & 0.62 \\
Molmo-O & 7B & 0.25 & 0.15 & 0.60 & 0.60 \\
Pixtral & 12B & 0.73 & 0.48 & 0.93 & 0.92 \\
\bottomrule
\end{tabular}
\caption{Table and chart extraction performance comparison across different models. Tables from PubTables and FinTabNet are extracted into HTML format and evaluated with the TEDS metric. Charts, from the ChartQA dataset, are extracted to either MD or HTML based on the prompt and evaluated with the mTEDS metric.}
\label{tab:model_comparison_chart_extraction}
\end{table}

\subsection{Safety Benchmarks}

To evaluate the safety capabilities of our Granite Vision model, we employ two different setups. First, we utilize the standard VLM-as-a-Judge setup as described in \citep{Li2024RedTV}. Second, we also introduce a new safety classification setup, where we formalize a range of safety tasks as classification problems. This new setup is aimed at building safer and more reliable AI models by leveraging strong classification capabilities. While existing generative MLLMs typically do not excel at classification tasks~\citep{mitra2024sparse,Huang2024MultimodalTV}, we believe that enhancing their discriminative capabilities is essential for enabling diverse applications and more sophisticated reasoning.


For the new safety classification setup, we also apply our Safety Vectors (SVs) approach (See Section~\ref{sec:model:sav}) to our Granite Vision model. Our method is compared with other strong baselines on three public benchmarks: VLGuard~\citep{zong2024safety},  RTVLM~\citep{Li2024RedTV}, and LMM-Halucination~\citep{chen2024unified}.



\paragraph{Benchmarks.} (i) \textbf{VLGuard}~\citep{zong2024safety} focuses on vision-language safety and identifies four main categories of harmful content: Privacy, Risky Behavior, Deception and Hateful Speech. The dataset consists of images from diverse sources and the instructions are generated by GPT-4V~\citep{OpenAI2023GPT4TR} with each safe image having both safe and unsafe instructions, and each unsafe image having a single instruction. (ii) \textbf{RTVLM}~\citep{Li2024RedTV} is the first red teaming dataset to benchmark current MLLMs in terms of several different aspects: faithfulness, privacy, safety, and fairness. In this dataset, 5,200 samples have been annotated by humans, or generated by GPT-4 accompanied by examples provided by humans. To ensure that the test data is unique and has not been seen by any of the evaluated VLMs, the authors produced new question-image pairs from publicly available images or generated from diffusion. (iii) \textbf{LMM-Halucination}~\citep{chen2024unified} is a dataset that evaluates the hallucinations of the models when answering multi-modal tasks. We use the default evaluation method provided in the dataset to identify whether this scenario is ``hallucinating'' or ``not hallucinating'', and compute the accuracy rate on correctly identified scenarios.

\paragraph{Baselines.} We compared Granite Vision's safety performance against the following MLLMs: (i) Phi-3.5-vision~\citep{abdin2024phi} is part of a family of small language models, which are designed for high capability and cost-effectiveness, performing well on tasks like language understanding, reasoning, coding, and math; (ii) LLaVA-v1.5-7B~\citep{liu2023llava} is a state-of-the-art MLLM that maps CLIP visual features to the LLM's embedding space and uses instruction tuning on the diverse LLaVA-Instruct-158k dataset for visual alignment. This dataset combines images with various response types, including conversational, descriptive, and reasoning, to enhance alignment; (iii) SmolVLM~\citep{smolVLM} is a new family of 2B small vision-language models capable of answering questions about images, describing visual content, creating stories based on multiple images, or functioning solely as a language model. Its lightweight design makes it ideal for on-device applications while ensuring strong performance.

We next discuss the two distinct safety evaluation setups.

\input{Tables/safety-table2}

\paragraph{VLM-as-a-Judge Setup.} Here, we perform a standard evaluation using the approach of VLM-as-a-Judge in the same manner as described in the RTVLM evaluation procedure (see Table 8 in~\cite{Li2024RedTV}). Particularly, we utilize GPT-4V, and the score has a range between [0,10] for both VLGuard and RTVLM benchmarks. In the evaluation of VLGuard, we distinguish between two categories of image-instruction pairs as provided by the original categories: ``Unsafe'', where both the images and instructions are unsafe, and ``Safe-Unsafe'', where the images are unsafe but the instructions are safe. For RTVLM, we evaluate using the same setup as in~\citep{Li2024RedTV}, focusing on four safety categories: `Mislead', `Politic', `Racial', and `Jailbreak'. Last, we compare with other baselines, like Phi-3.5-vision~\citep{abdin2024phi}, LLaVA-v1.5-7B~\citep{liu2023llava}, SmolVLM~\citep{smolVLM}.




Results are shown in Table~\ref{tbl:safety}. 
As can be seen, Granite Vision performs comparably to other baselines on the standard safety evaluation setup. Readers must note that Granite Vision is the most lightweight model compared to baselines. We believe that there's definitely scope for further improvement and we discuss this in our future work. In the next section, we describe our new efforts to formalize a wide range of safety tasks as classification problems.

\paragraph{Safety Classification Setup.} In this setup, we reformat the tasks in VLGuard to be a classification task and we report the raw classification accuracy of our method. Specifically, we prompt the model with labels, and ask it to select the right answer, comparing it to the ground truth. The motivation behind reformulating safety tasks as classification problems is that it encourages the development of MLLMs with robust discriminative abilities, crucial for safety-critical applications, and provides a clear, quantifiable metric for assessing safety performance. Last, we also apply our Safety Vectors (SVs) approach (See Section~\ref{sec:model:sav}) to our Granite Vision model.

For VLGuard, the set of class labels is given by $\mathcal{C} = \{\text{``safe"}, \text{``unsafe"}\}$. We use the metric VLGuard proposed Attack Success Rate (ASR) for evaluating unsafe inputs and Helpfulness for evaluating safe inputs. Note that accuracy is the complement of the ASR and can be calculated as 1 - ASR. The prompt we use to query the model is ``Is the image and instruction harmful? Answer the question with Yes or No." In our implementation, we extract the SVs from the train set and evaluate on the test set. In this experiment, we focus solely on the ``Unsafe" category, as it is the most relevant aspect for safety evaluation. For MHalu, the set of class labels for this task is given by $\mathcal{C} = \{\text{``hallucinating"}, \text{``not hallucinating"}\}$, and the prompt we use to query the model is ``Is the Claim hallucinating? Answer the question with Yes or No."


\begin{wraptable}{r}{0.4\textwidth}
    \vspace{-10pt}
    \begin{tabular}{lccc}
    \toprule
    & \multicolumn{1}{c}{MHalu} & \multicolumn{1}{c}{VLGuard} & \\
    \midrule
    
    Granite Vision & 78.0 & 86.0  \\
    \hline
    \textbf{+Safety Vectors} & {80.7} & {96.2}  \\
    \bottomrule
    \end{tabular}
    \vspace{-3pt}
    \caption{\textit{Safety classification} {Results}.}
    \label{tab:vlm-judge}
\end{wraptable}


Results are shown in Table~\ref{tab:vlm-judge}. 
It can be observed from the table that Granite Vision performs relatively well in terms of safety when the task is reformulated as a classification task. This is somewhat surprising, as MLLMs typically struggle with discriminative vision-and-language tasks~\citep{mitra2024sparse}, and safety classification poses a significant challenge. Additionally, our approach for finding SVs seems to be more effective, probably due to the fact that this approach can utilize multimodal features for downstream discriminative tasks. Overall, we believe that framing safety evaluation as a classification task may offer a valuable framework for improving the safety of AI models, and we anticipate that future research on this aspect will provide further advancements in safety research.

\paragraph{Summary and Future Directions.} Ensuring the safety of generative MLLMs is absolutely crucial in order to prevent harm, build trust, address ethical concerns, and enable their responsible deployment in real-world applications. Our results demonstrate that Granite Vision performs almost at par with baselines (despite being the lightest MLLM in the comparison pool) for {\it VLM-as-a-Judge} task. Notably, the addition of \textbf{Safety Vectors} to Granite Vision leads to a significant improvement in safety classification performance. We do acknowledge that further work needs to be done to improve high-level reasoning and correct occasional incorrect outputs to improve reliability in sensitive tasks, which require nuanced classification. To address these, we will incorporate more reasoning-focused and structure-related data into the training process in the future.

In addition, we showed in this paper that finding safety vectors (SVs) in Granite Vision's attention heads led to significant improvements when safety tasks were reformulated as classification problems. Current reliance for SVs is on few-shot samples which are informative but may have limited scope in terms of capturing the range of possible safety issues that can be encountered. To further improve the model's ability to identify and address all safety concerns, we plan to investigate scaling up SVs using more training data in future research.



\section{Conclusion} 
In this paper, we presented Granite Vision, a compact large language model with integrated vision capabilities, tailored to meet the unique demands of enterprise use cases, in particular visual document understanding. Our model achieves strong results in standard benchmarks and is publicly available under a permissive license for both research and commercial use. Our future work includes enabling multi-page document processing through context compression, learning with structured inputs and outputs, multi-hop reasoning, steering model activations for enhanced reasoning capabilities, and exploring techniques to preserve Granite Vision language-only capabilities.

\bibliography{references}
\bibliographystyle{iclr2025_conference}

\clearpage

\section{Appendix}

\begin{longtable}{@{} l r r @{}}
\caption{A comprehensive overview of document understanding datasets used in Granite Vision.} \label{tab:doc_datasets}\\
\toprule
\textbf{Datasets} & \textbf{\# Images} & \textbf{\# QA Pairs} \\
\midrule
\endfirsthead

\multicolumn{3}{c}{\tablename\ \thetable\ -- \textit{Continued from previous page}} \\
\toprule
\textbf{Datasets} & \textbf{\# Images} & \textbf{\# QA Pairs} \\
\midrule
\endhead

\midrule \multicolumn{3}{r}{\textit{Continued on next page}} \\ 
\endfoot

\bottomrule
\endlastfoot

\multicolumn{3}{@{}l}{\textbf{Document QA}} \\
\midrule
\textcolor{Cyan40}{DocFM-VQA} & 2.4 M   & 19.9 M \\
\textcolor{Purple50}{Docmatix} \citep{laurençon2024building} & 565.0 K & 3.9 M \\
\textcolor{Purple50}{synthdog-en} \citep{kim2022ocr} & 500.0 K & 500.0 K \\
\textcolor{Purple50}{pixmo-docs} \citep{deitke2024molmo} & 255.4 K & 2.3 M \\
\textcolor{Purple50}{HME100K} \citep{yuan2022syntax} & 74.5 K  & 74.5 K \\
\textcolor{Purple50}{Ureader QA} \citep{ye2023ureader} & 72.7 K  & 370 K \\
\textcolor{Purple50}{ArxivQA} \citep{li2024multimodal} & 54.4 K  & 100 K \\
\textcolor{Purple50}{Ureader KG} \citep{ye2023ureader} & 37.6 K  & 37.6 K \\
\textcolor{Cyan40}{Tech Book QA} & 10.4 K  & 17.8 K \\
\textcolor{Purple50}{DocVQA} \citep{mathew2021docvqa} & 10.2 K  & 39.5 K \\
\textcolor{Purple50}{IAM} \citep{marti2002iam} & 5.7 K   & 5.7 K \\
\textcolor{Cyan40}{Business Document QA} & 3.3 K   & 7.8 K \\
\textcolor{Green40}{Visualmrc} \citep{tanaka2021visualmrc} & 3.0 K   & 12 K \\
\midrule
\multicolumn{3}{@{}l}{\textbf{Chart QA}} \\
\midrule
\textcolor{Purple50}{Unichart} \citep{masry2023unichart} & 611.0 K & 6.9 M \\
\textcolor{Purple50}{Tinychart} \citep{zhang2024tinychart} & 606.0 K & 4.5 M \\
\textcolor{Cyan40}{DocFM-VQA-Charts} & 548.7 K & 5.5 M \\
\textcolor{Purple50}{Dvqa} \citep{kafle2018dvqa} & 200.0 K & 2.3 M \\
\textcolor{Green40}{PlotQA} \citep{methani2020plotqa} & 157.0 K & 20.2 M \\
\textcolor{Green40}{FigureQA} \citep{kahou2017figureqa} & 100.0 K & 1.3 M \\
\textcolor{Cyan40}{DocFM-VQA-AugChartQA} & 35.1 K  & 294.2 K \\
\textcolor{Purple50}{Chart2Text} \citep{obeid2020chart} & 27.0 K  & 30.0 K \\
\textcolor{Green40}{ChartQA} \citep{masry2022chartqa} & 18.3 K  & 28.3 K \\
\textcolor{Purple50}{Vistext} \citep{tang2023vistext} & 9.9 K   & 9.9 K \\
\midrule
\multicolumn{3}{@{}l}{\textbf{Infographics QA}} \\
\midrule
\textcolor{Purple50}{Infographic VQA} \citep{mathew2022infographicvqa} & 8.5 K & 35.7 K \\
\midrule
\multicolumn{3}{@{}l}{\textbf{Table QA}} \\
\midrule
\textcolor{Purple50}{RoBUT-wikiSQL} \citep{zhao2023robut} & 75.0 K  & 86.2 K \\
\textcolor{Purple50}{RoBUT-WTQ} \citep{zhao2023robut} & 38.0 K  & 44.1 K \\
\textcolor{Purple50}{tabmwp} \citep{lu2022dynamic} & 22.7 K  & 23 K \\
\textcolor{Purple50}{RoBUT-SQA} \citep{zhao2023robut} & 8.5 K   & 34.1 K \\
\textcolor{Purple50}{MultiHiertt} \citep{zhao2022multihiertt} & 7.6 K   & 7.8 K \\
\textcolor{Purple50}{Finqa} \citep{chen2021finqa} & 5.3 K   & 6.25 K \\
\textcolor{Purple50}{Hitab} \citep{cheng2022hitab} & 2.5 K   & 7.8 K \\
\midrule
\multicolumn{3}{@{}l}{\textbf{Diagram QA}} \\
\midrule
\textcolor{Purple50}{TQA} \citep{kembhavi2017you} & 27.3 K  & 29.8 K \\
\textcolor{Cyan40}{DocFM-VQA-Flowcharts} & 16.6 K  & 74.6 K \\
\textcolor{Purple50}{AI2D GPT4V} \citep{Li2024LLaVAOneVisionEV} & 4.9 K & 4.9 K \\
\textcolor{Purple50}{AI2D InternVL} \citep{Li2024LLaVAOneVisionEV} & 3.2 K & 12.4 K \\
\textcolor{Purple50}{AI2D} \citep{kembhavi2016diagram} & 2.4 K   & 7.5 K \\
\textcolor{Purple50}{Diagram Image2Text}\footnotemark[7] & 0.3 K   & 0.3 K \\
\midrule
\multicolumn{3}{@{}l}{\textbf{Reasoning / Captioning}} \\
\midrule
\textcolor{Cyan40}{DocFM Visual Cue \& Captioning} & 1.2 M   & 2.1 M \\
\textcolor{Purple50}{Textcaps} \citep{sidorov2020textcaps} & 22.0 K  & 22.0 K \\
\textcolor{Purple50}{DocReason}\footnotemark[8] & 8.7 K   & 25.8 K \\
\midrule
\multicolumn{3}{@{}l}{\textbf{Grounding / Text Localization}} \\
\midrule
\textcolor{Green40}{Multi-grained Text Localization} \citep{hu2024mplug} & 1.0 M   & 1.0 M \\
\textcolor{Cyan40}{DocFM Rule Based Grounding} & 190 K   & 534.4 K \\
\textcolor{Cyan40}{KVP10K} \citep{naparstek2024kvp10k} & 8.0 K   & 8.0 K \\
\midrule
\multicolumn{3}{@{}l}{\textbf{Scene Text}} \\
\midrule
\textcolor{Purple50}{K12 Printing} \citep{li2024llava} & 256.6 K & 256.6 K \\
\textcolor{Purple50}{OCR-VQA} \citep{mishra2019ocr} & 165.7 K & 801.6 K \\
\textcolor{Purple50}{TextOCR} \citep{singh2021textocr} & 25.1 K  & 25.1 K \\
\textcolor{Purple50}{Ureader CAP} \citep{ye2023ureader} & 22.0 K  & 91.4 K \\
\textcolor{Purple50}{TextVQA} \citep{singh2019towards} & 22.0 K  & 34.6 K \\
\textcolor{Purple50}{Llavar} \citep{zhang2023llavar} & 19.8 K  & 43.2 K \\
\textcolor{Purple50}{ST-VQA} \citep{biten2019scene} & 16.4 K  & 22 K \\
\textcolor{Purple50}{RenderedText}\footnotemark[9] & 10.0 K  & 10.0 K \\
\textcolor{Purple50}{IIIT5K} \citep{mishra2012scene} & 2.0 K   & 2.0 K \\
\midrule
\multicolumn{3}{@{}l}{\textbf{UI Screen / Code}} \\
\midrule
\textcolor{Purple50}{Datikz-v2}\footnotemark[10] & 94.4 K  & 94.4 K \\
\textcolor{Purple50}{Datikz} \citep{belouadi2023automatikz} & 47.9 K  & 48.2 K \\
\textcolor{Purple50}{screenqa} \citep{hsiao2022screenqa} & 33.2 K  & 80.8 K \\
\textcolor{Purple50}{screen2words} \citep{wang2021screen2words} & 15.7 K  & 15.7 K \\
\textcolor{Purple50}{websight} \citep{laurenccon2024unlocking} & 10.0 K  & 10.0 K \\
\textcolor{Purple50}{Omniact} \citep{kapoor2024omniact} & 4.9 K   & 0.13 K \\
\midrule
\multicolumn{3}{@{}l}{\textbf{Structure Parsing}} \\
\midrule
\textcolor{Green40}{OCR-CC} \citep{yang2021tap} & 1.0 M   & 1.0 M \\
\textcolor{Green40}{CCpdf} \citep{turski2023ccpdf} & 937.8 K & 937.8 K \\
\textcolor{Purple50}{PubTables} \citep{smock2022pubtables} & 569.3 K & 569.3 K \\
\textcolor{Cyan40}{Chart2Table} & 550.4 K & 550.4 K \\
\textcolor{Green40}{TURL} \citep{deng2022turl} & 200.0 K & 200.0 K \\
\textcolor{Cyan40}{PubTabNet} \citep{zhong2020image} & 199.8 K & 199.8 K \\
\textcolor{Green40}{RVL-CDIP} \citep{harley2015evaluation} & 159.4 K & 159.4 K \\
\textcolor{Green40}{DUE} \citep{borchmann2021due} & 104.5 K & 1.6 M \\
\textcolor{Cyan40}{FinTabNet} \citep{zheng2020global} & 98.9 K  & 98.9 K \\
\end{longtable}

\footnotetext[7]{\url{https://huggingface.co/datasets/Kamizuru00/diagram_image_to_text}}
\footnotetext[8]{\url{https://huggingface.co/datasets/mPLUG/DocReason25K}}
\footnotetext[9]{\url{https://huggingface.co/datasets/wendlerc/RenderedText}}
\footnotetext[10]{\url{https://huggingface.co/datasets/nllg/datikz-v2}}

\newpage

\section{Contributions And Acknowledgments}

Authors (alphabetical order):

\vspace{0.15in}
{\bf Granite Vision Technical Leadership:} 

Assaf Arbelle,
Leonid Karlinsky,
Peter Staar,
Rogerio Feris,
Tal Drory

\vspace{0.15in}
{\bf Project Management:}

Abraham Daniels

\vspace{0.15in}
{\bf Core Contributors:}

Ahmed Nassar,
Amit Alfassi,
Bo Wu,
Eli Schwartz,
Dhiraj Joshi,
Jovana Kondic,
Nimrod Shabtay,
Pengyuan Li,
Roei Herzig,
Shafiq Abedin,
Shaked Perek,
Sivan Harary,
Udi Barzelay

\vspace{0.15in}
{\bf Contributors:}

Adi Raz Goldfarb,
Aude Oliva,
Ben Wieles,
Bishwaranjan Bhattacharjee,
Brandon Huang,
Christoph Auer,
Dan Gutfreund,
David Beymer,
David Wood,
Hilde Kuehne,
Jacob Hansen,
Joseph Shtok,
Ken Wong,
Luis Angel Bathen,
Mayank Mishra,
Maksym Lysak,
Michele Dolfi,
Mikhail Yurochkin,
Nikolaos Livathinos,
Nimrod Harel,
Ophir Azulai,
Oshri Naparstek,
Rafael Teixeira de Lima,
Rameswar Panda,
Sivan Doveh,
Shubham Gupta,
Subhro Das,
Syed Zawad,
Yusik Kim,
Zexue He

\vspace{0.15in}
{\bf Platform: }

Alexander Brooks,
Gabe Goodhart

\vspace{0.15in}
{\bf Product Management:}

Anita Govindjee,
Derek Leist,
Ibrahim Ibrahim

\vspace{0.15in}
{\bf IBM Research Leadership:}

Aya Soffer,
David Cox,
Kate Soule,
Luis Lastras,
Nirmit Desai,
Shila Ofek-koifman,
Sriram Raghavan,
Tanveer Syeda-Mahmood

\vspace{0.15in}
{\bf Acknowledgments:}

We would like to acknowledge Amy Angelini, Bob Calio, Brian Belgodere, Dakshi Agrawal, Heiko Ludwig, Hendrik Strobelt, John Smith, Kim Martineau, Kush Varshney, Mauro Martino, Petros Zerfos, Ray Rose, and Sandeep Gopisetty. We would also like to acknowledge the support from IBM Research AI and the MIT-IBM Watson AI lab.

\end{document}

%% file: Tables/safety-table2.tex
\newcolumntype{N}{>{\centering\arraybackslash}p{1cm}} 
\newcolumntype{C}{>{\centering\arraybackslash}X}   

\renewcommand{\arraystretch}{1.0}
\setlength{\tabcolsep}{3pt} 
\begin{table*}[t]
\centering
\small
\begin{tabularx}{\textwidth}{
    >{\raggedright\arraybackslash}p{2.9cm} 
    *{9}{>{\centering\arraybackslash}X} 
    >{\centering\arraybackslash}p{1.5cm} 
}
\toprule
\textbf{Model} &
\multicolumn{2}{c}{\textbf{VLGuard}} &
\multicolumn{4}{c}{\textbf{RTVLM}}
\\
\cmidrule(lr){2-3}
\cmidrule(lr){4-7}
& \multicolumn{1}{c}{Unsafe} & \multicolumn{1}{c}{Safe-Unsafe} & 
\multicolumn{1}{c}{Mislead} & \multicolumn{1}{c}{Politics} & \multicolumn{1}{c}{Racial} & \multicolumn{1}{c}{Jailbreak}
\\
\toprule

    LLaVA-v1.5-7B & {5.3} & {7.4} & {8.6} & {7.3} & {7.2} & {4.4}  \\
    Phi3.5-vision & {8.7} & {9.3} & {8.5} & {8.2} & {8.2} & {9.3}  \\
    SmolVLM & {4.8} & {7.7} & {5.6} & {6.0} & {4.6} & {7.3}  \\

   
    \hline
    Granite Vision (Ours) & 6.5 & 8.6 & {7.7} & {7.2} & {7.7} & {4.5}  \\
\bottomrule
\end{tabularx}
\caption{\textbf{Results} for the \textit{VLM-as-a-Judge} setup. We evaluate on the VLGuard and RTVLM benchmarks. The score has a range between [0,10], and higher is better.
}
\label{tbl:safety}
\end{table*}